\documentclass[11pt]{article}

\usepackage[]{acl}
\usepackage{times}
\usepackage{latexsym}

\usepackage[T1]{fontenc}
\usepackage[utf8]{inputenc}
\usepackage{microtype}
\usepackage{graphicx}

\newcommand{\hide}[1]{}
\newcommand{\vpara}[1]{\vspace{0.05in}\noindent \textbf{#1 }}
\usepackage{amssymb}
\usepackage{booktabs}
\usepackage{multirow}
\usepackage{amsmath}
\usepackage{amsfonts}
\usepackage{hyperref}
\usepackage{url}
\usepackage{xcolor}
\usepackage{tikz}
\usetikzlibrary{shapes,arrows,positioning,calc}
\usepackage{tcolorbox}
\tcbuselibrary{skins,breakable}
\usepackage{algorithm}
\usepackage[noend]{algpseudocode}
\algrenewcommand{\algorithmiccomment}[1]{\hfill$\triangleright$\; #1}

\title{ActiveMem: Distributed Active Memory for Long-Horizon LLM Reasoning}

\author{Yunhan Jiang\textsuperscript{1, 2}, Wenbin Duan\textsuperscript{1, 2}, Shasha Guo\textsuperscript{1}\thanks{Corresponding authors.}, Liang Pang\textsuperscript{1}\footnotemark[1],\\
\textbf{Xiaoqian Sun\textsuperscript{1}}, \textbf{Huawei Shen\textsuperscript{1}} \\
  \textsuperscript{1}State Key Laboratory of AI Safety, Institute of Computing Technology, \\Chinese Academy of Sciences\\
  \textsuperscript{2}University of Chinese Academy of Sciences\\
  jiangyunhan20@mails.ucas.ac.cn \\
  \{duanwenbin25e, guoshasha, pangliang, sunxiaoqian, shenhuawei\}@ict.ac.cn
  }

\begin{document}
\maketitle

\begin{abstract}
Memory is essential for enabling large language model (LLM) agents to handle long-horizon reasoning tasks. Existing memory mechanisms are largely centralized, typically organizing retrieved information and interaction history within a single model context. 
This design imposes a fundamental trade-off: scaling reasoning trajectories risks context overload, whereas aggressive content pruning may result in irreversible information loss.
Seeking a better trade-off, we draw inspiration from human cognitive systems, especially the functional complementarity between the prefrontal cortex (executive control) and the hippocampus (memory management), suggesting that such a trade-off need not be inherent, but may instead stem from centralized memory organization. To this end, we propose ActiveMem, a heterogeneous framework that decouples agent memory from the core reasoning process. 
Specifically, a high-level Planner utilizes distilled semantic gists to execute reasoning, while a lightweight, distributed memory system operates in parallel to actively accumulate and consolidate these gists throughout the task.
Experiments on BrowseComp-Plus and GAIA show that ActiveMem achieves state-of-the-art accuracy with significantly reduced overhead, demonstrating the effectiveness of distributed active memory for long-horizon reasoning.

\end{abstract}

\section{Introduction}
\label{sec:introduction}

LLM agents have demonstrated remarkable capabilities in executing long-horizon reasoning tasks through sustained, multi-step interactions~\citep{yao2023react, nakano2021webgpt, wang2024survey}. 
However, within these intricate workflows, the continuous expansion of interaction contexts inevitably renders working memory management a critical bottleneck~\citep{zhang2025survey, DBLP:journals/corr/abs-2512-13564}. 
To this end, an effective working memory must selectively retain task-relevant information and anchor the model's attention on pivotal tokens while compressing the active context window, thereby enabling agents to navigate complex, long-horizon tasks successfully.

\begin{figure}[!t]
    \centering
\includegraphics[width=\columnwidth]{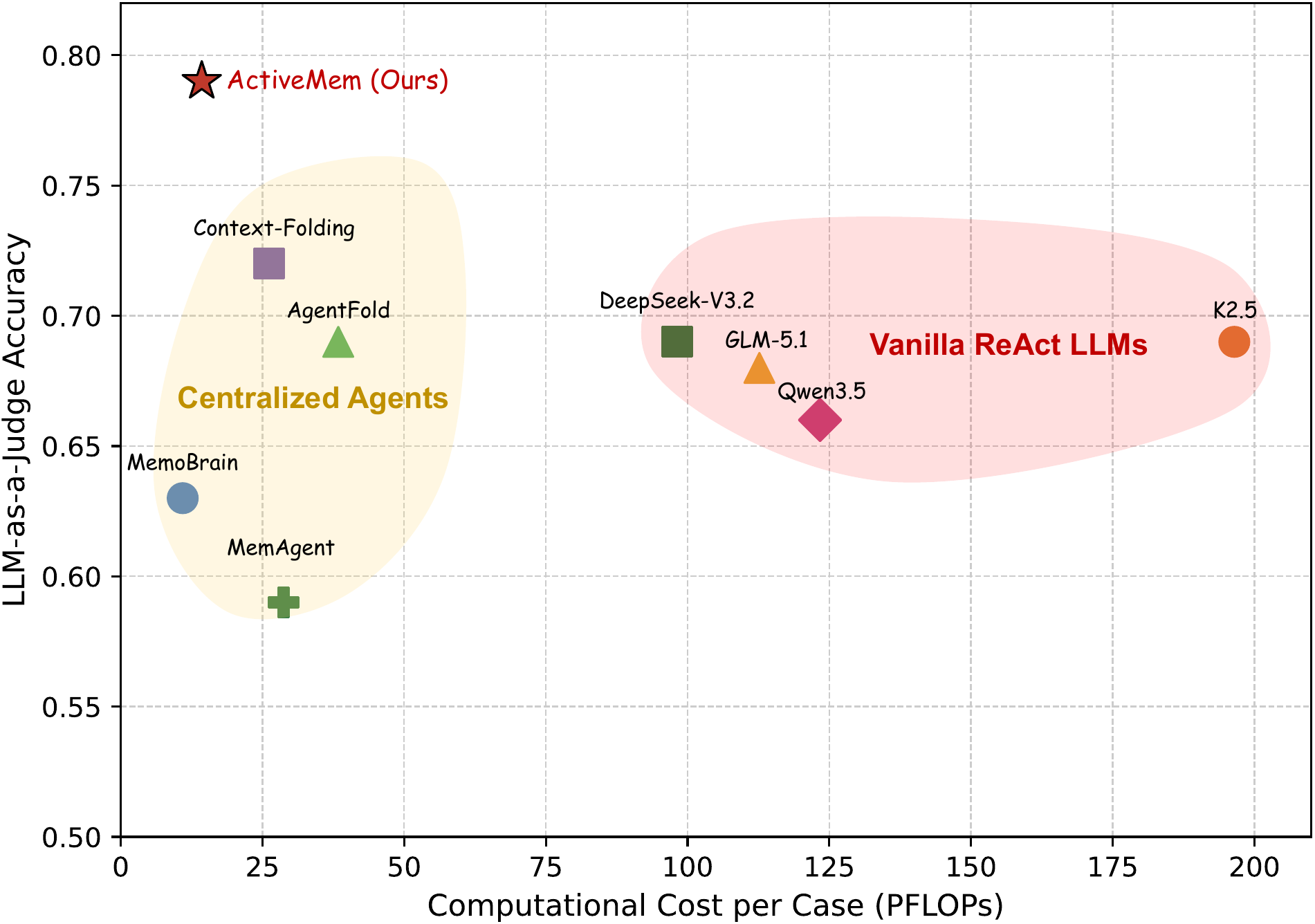}
    \caption{ActiveMem outperforms both modern centralized memory agents and vanilla ReAct LLMs in LLM-as-a-Judge accuracy while achieving substantially lower computational cost.}
    \label{fig:pflops_acc}
    \vspace{-15pt}
\end{figure}

% ==================== 插入图 1 ====================

\begin{figure*}[!t]
    \centering
    \includegraphics[width=0.95\textwidth]{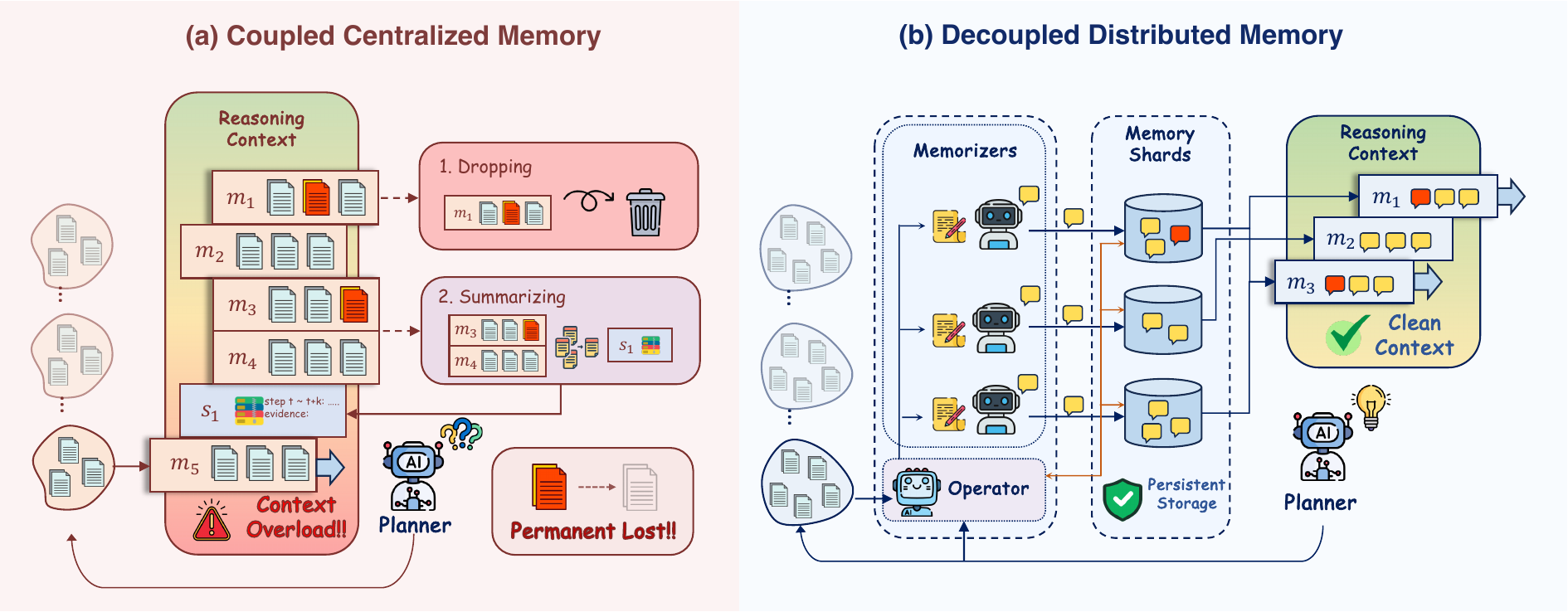}
    \caption{Comparison between (a) \textbf{Coupled Centralized Memory} and (b) our proposed \textbf{Decoupled Distributed Memory} (ActiveMem). In the centralized paradigm, existing approaches manage context growth by selectively retaining memories or compressing them into step-level summaries, trading information completeness for a bounded context. ActiveMem takes a different path: evidence is routed to parallel Memorizers that produce distilled semantic gists, stored persistently in Memory Shards and coordinated by an Operator, enabling the Planner to reason over a consistently compact context without discarding the underlying information.}
    \label{fig:teaser}
    
\end{figure*}

% ====================================================

Despite this necessity, most current reasoning systems follow a centralized architecture where memory is tightly bound to a single core reasoner. In ReAct-style agents, for instance, retrieved information and intermediate trajectories continuously accumulate within the same model context window~\citep{yao2023react}. As reasoning chains extend, this centralization inevitably triggers severe context overload~\citep{DBLP:conf/acl/LevyJG24, DBLP:conf/nips/AnML0LC24} and the \textit{lost-in-the-middle} phenomenon~\citep{liu2023lost, DBLP:conf/icml/ShiCMSDCSZ23}, thereby undermining reasoning performance. To mitigate this issue, contemporary approaches introduce various context compression mechanisms~\citep{sun2025scaling, ye2025agentfold, zhu2025memobrain, DBLP:journals/corr/abs-2509-13313, DBLP:journals/corr/abs-2506-15841, DBLP:journals/corr/abs-2510-12635, DBLP:journals/corr/abs-2507-13334}. However, these strategies invariably incur permanent information loss--- \textit{either by dropping older memories entirely or compressing them into coarse step-level summaries}. This leaves memory content irreversibly degraded and unrecoverable for subsequent reasoning steps (Figure~\ref{fig:teaser} (a)). This dilemma exposes a fundamental limitation of centralized memory designs: memory storage and reasoning computation are tightly coupled, creating an inherent trade-off between trajectory scaling and memory fidelity.

% Drawing a functional analogy from cognitive neuroscience, we leverage the complementary organization of the prefrontal cortex (PFC) and the hippocampus.
% In human cognition, the PFC issues top-down executive signals to guide retrieval from external memory systems rather than serving as the primary repository for detailed memory content~\citep{lara2015role}. The hippocampus then executes pattern completion, integrating these executive signals to reactivate relevant information distributed across the neocortex.  This reactivation process is notably holistic----co-reactivating memory elements across multiple neocortical regions simultaneously~\citep{horner2015evidence}---and abstract, frequently trading granular episodic details for structured semantic gist ~\citep{hindy2026hippocampal}. 
% This biological mechanism suggests a promising architectural principle for LLM agents: structurally decoupling memory systems from reasoning processes. Under this paradigm, the central reasoner is reserved for high-level reasoning over distilled evidence, while memory persistence and retrieval are delegated to a dedicated, distributed subsystem. 
% Furthermore, analogous to the parallel nature of hippocampal pattern completion, evidence processing within agent workflows is inherently parallelizable. Multiple lightweight distributed modules can simultaneously process segmented evidence streams to generate distilled semantic gists, which subsequently persist and consolidate within local memory shards across the task lifecycle. 

Drawing inspiration from human cognitive systems, we argue that this seemingly inherent trade-off stems fundamentally from the limitations of centralized memory organization. The human brain masterfully circumvents this bottleneck through the functional complementarity between the prefrontal cortex (executive control) and the hippocampus (memory management). The prefrontal cortex (PFC), serving as the master executive controller, issues top-down executive signals to guide retrieval rather than acting as a massive repository for detailed memory content~\citep{lara2015role}. Complementarily, the hippocampus executes parallel pattern completion, integrating these executive signals to reactivate and distribute holistic, abstract information across the neocortex~\citep{horner2015evidence}, frequently trading granular episodic details for structured distilled semantic gists~\citep{hindy2026hippocampal}. This biological mechanism suggests a promising design paradigm: structurally decoupling memory systems from reasoning processes.
% Under this decoupled paradigm, the central reasoner is exclusively reserved for high-level reasoning over distilled gists, while memory persistence and retrieval are delegated to a dedicated, distributed subsystem. Crucially, analogous to the parallel nature of hippocampal pattern completion, information processing within agent workflows becomes inherently parallelizable, enabling multiple lightweight modules to concurrently process segmented streams, which subsequently persist and consolidate within local memory shards across the task lifecycle.

% Motivated by these biological insights, we introduce \textbf{ActiveMem}, a heterogeneous framework that decouples memory management from high-level reasoning. ActiveMem consists of two parts: a \textit{Planner} that handles top-down reasoning and query generation, and a \textit{Distributed Memory System} comprising three components---\textit{Memorizers} that process retrieved documents in parallel to extract distilled semantic gists, \textit{Memory Shards} that persistently store and consolidate these gists throughout the task, and an \textit{Operator} that coordinates routing and reuse across shards. As shown in Figure~\ref{fig:teaser}(b), this design keeps the Planner's context compact and bounded while preserving document-level information, substantially mitigating the dilemma between context overload and information loss inherent to centralized memory paradigms.

Motivated by the above insight, we introduce \textbf{ActiveMem}, a heterogeneous framework that implements a decoupled and distributed active memory architecture to overcome the limitations of centralized paradigms. Specifically, ActiveMem consists of two primary modules: a high-level \textit{Planner} and a \textit{Distributed Memory System}. The Planner handles reasoning and top-down query generation, remaining exclusively focused on executing core reasoning chains over compact context windows. Complementing this, the Distributed Memory System replaces monolithic context buffers with a parallelized and sharded architecture that is inherently lightweight and active. This architecture comprises three tightly coordinated components:
(1) \textit{Memorizers}, which exploit the inherent parallelizability of information processing to concurrently process retrieved documents and extract distilled semantic gists;
(2) \textit{Memory Shards}, which actively partition, persistently store, and consolidate these gists across localized nodes throughout the task lifecycle; and
(3) an \textit{Operator}, which dynamically orchestrates proactive routing and semantic reuse across the entire shard network. 
As illustrated in Figure~\ref{fig:teaser}(b), this collaborative design keeps the Planner's context horizon bounded and clean while preserving document-level insights, thereby substantially mitigating the trade-off between trajectory scaling and memory fidelity.

\textbf{Contributions.} 
(1)\textbf{Neuroscience-inspired cognitive decoupling.} We propose a decoupled memory-reasoning paradigm inspired by the functional synergy between the prefrontal cortex and hippocampus. This architecture frees the centralized reasoning core by segregating executive control from distributed memory consolidation.
(2) \textbf{The ActiveMem framework.} 
We introduce ActiveMem, a heterogeneous framework that materializes this paradigm into a distributed and active memory architecture. It empowers Memorizers to concurrently process retrieved documents and synthesize consolidated semantic gists, which are dynamically maintained across localized shards throughout the task lifecycle.
(3) \textbf{Superior accuracy with lower computational cost.}
ActiveMem achieves the highest LasJ accuracy among nine baselines while incurring substantially lower computational complexity in terms of PFLOPs, as shown in Figure~\ref{fig:pflops_acc}.

% In summary, the main contributions of this paper are as follows:
% \begin{itemize}
%     \item \textbf{Neuroscience-inspired perspective}: We draw on the PFC--hippocampus relationship as a computational analogy that motivates a decoupled design where a central planner issues retrieval guidance while distributed modules reactivate and distill task-relevant information.
%     \item \textbf{Novel architecture}: We propose ActiveMem, a heterogeneous framework in which lightweight Memorizers process localized evidence shards in parallel and return distilled semantic gists to the Planner via persistent Memory Shards.
%     \item \textbf{Superior empirical performance}: ActiveMem achieves the highest LLM-as-a-Judge accuracy among nine baselines on long-horizon benchmarks while requiring substantially less computation, as shown in Figure~\ref{fig:pflops_acc}.
% \end{itemize}
\section{Related Work}
\label{sec:related}

Memory management has become a central challenge in LLM agents. Existing centralized approaches can be grouped into three forms. The first is vanilla centralized memory, where raw trajectories or retrieved documents are directly fed back into one model context, as in ReAct-style agents~\citep{yao2023react}. The second targets long-term dialogue management---methods such as A-MEM~\citep{wang2025amem}, Mem0~\citep{chhikara2025mem0}, MemGPT~\citep{DBLP:journals/corr/abs-2310-08560}, MemoryBank~\citep{DBLP:conf/aaai/ZhongGGYW24}, Memory-R1~\citep{DBLP:journals/corr/abs-2508-19828}, and Mem-$\alpha$~\citep{DBLP:journals/corr/abs-2509-25911} focus on preserving user interaction histories across extended conversations rather than supporting tool-augmented long-horizon reasoning. The third addresses long-horizon reasoning more directly, either through compression-based strategies that summarize past trajectories or raw documents~\citep{sun2025scaling, ye2025agentfold, yu2025memagent, DBLP:journals/corr/abs-2510-12635}, or through structured organization over working memory~\citep{hu2025hiagent, zhu2025memobrain}. These methods improve how agents manage working memory and advance the ability to handle long-horizon tasks, but memory and reasoning remain tightly coupled within one central reasoning node, creating a trade-off between retaining sufficient detail and maintaining efficient inference.

A smaller but growing line of work explores distributed memory across multiple agents or modules. MIRIX~\citep{wang2025mirix} introduces a modular multi-agent memory system that assigns memory consolidation to specialized controllers organized by memory type, such as episodic, semantic, and procedural memory. However, this form of distribution follows a memory taxonomy rather than the evolving information needs of the reasoning process. Its memory modules therefore do not actively distill task-relevant information under Planner-issued sub-queries, limiting their effectiveness for long-horizon reasoning tasks where relevant evidence must be dynamically interpreted and selectively surfaced to a central reasoner.

% Our work differs from both lines of research. Centralized methods keep memory and reasoning tightly coupled, and those that compress context to manage overload do so at the cost of irreversible information loss. Existing distributed methods, such as MIRIX, organize memory by type rather than by reasoning need, falling short for long-horizon tasks. ActiveMem fills this gap: parallel Memorizers distill raw evidence into persistent semantic gists under Planner-issued sub-queries, decoupling memory from reasoning while eliminating the permanent information loss inherent in centralized compression.

Our work differs from both lines of research. Centralized methods keep memory and reasoning tightly coupled within a single model context. Methods that compress or truncate this context can reduce overload, but may discard information needed for later reasoning. Existing distributed methods, like MIRIX, organize memory by memory type rather than by the information needs of the reasoning process, which limits their ability to support long-horizon tasks that require dynamic document selection. ActiveMem addresses this limitation by decoupling memory formation from reasoning. Parallel Memorizers process raw documents under queries issued by the Planner and produce query-conditioned memory summaries, which are stored in persistent Memory Shards and selectively returned to the Planner when needed.
\section{Method}
\label{sec:method}

% ActiveMem explicitly separates reasoning from memory formation. The reasoning system centers on a \textbf{Planner} that issues top-down retrieval signals and integrates distilled evidence for global reasoning. The memory system consists of distributed \textbf{Memory Shards} coordinated by an \textbf{Operator} and populated by lightweight parallel \textbf{Memorizers}. Figure~\ref{fig:framework} illustrates the overall architecture and its single-step ReAct cycle.

\begin{figure*}[t]
    \centering
    \includegraphics[width=\textwidth]{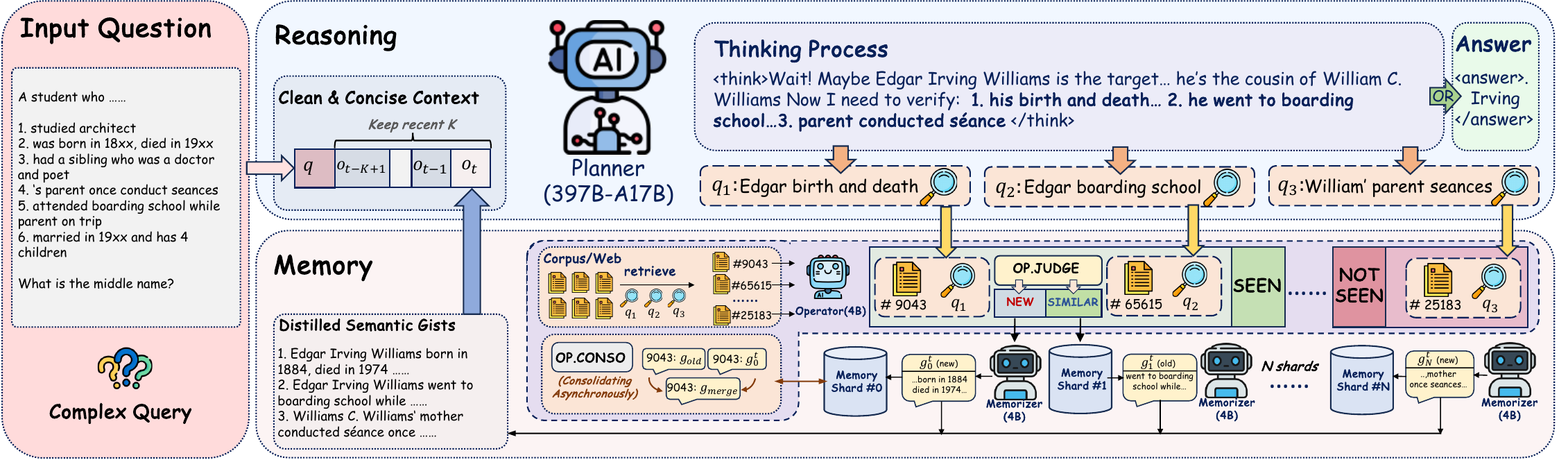}
    \caption{Overview of our ActiveMem framework. The Planner issues retrieval queries $\mathcal{Q}_t$ to recall documents from an external corpus. Each document is paired with its query to form a memory task and routed to the appropriate shard. For repeated documents, the Operator checks semantic similarity to a prior task: if similar, the stored gist is returned directly; otherwise, the Memorizer distills a new gist for the Planner while the Operator consolidates it into the shard asynchronously. This decouples memory from reasoning---the memory system continuously accumulates gists in persistent shards, while the Planner operates over a clean, compact distilled context.}
    \label{fig:framework}
    \vspace{-12pt}
\end{figure*}

We introduce ActiveMem, a distributed memory framework that decouples memory management from high-level reasoning. The Planner generates retrieval queries and integrates the returned semantic gists to guide subsequent reasoning or produce the final answer. The Distributed Memory System consists of persistent Memory Shards, lightweight Memorizers and Operator: Memorizers process retrieved documents in parallel and produce distilled semantic gists; Memory Shards actively accumulate and consolidate these gists throughout the task; and the Operator coordinates routing, memory reuse, and consolidation across shards. Figure~\ref{fig:framework} illustrates the overall architecture of ActiveMem.

\subsection{Planner}

% The Planner maintains a compact reasoning state $s_t = (x, h_t, m_t)$, where $x$ is the original question, $h_t$ is a trimmed recent interaction history, and $m_t$ is the distilled memory returned by the memory system at step $t$. 
The Planner maintains a compact reasoning state $s_t = (x, h_t, m_{t-1})$, where $x$ is the original question, $h_t$ records the trimmed interaction history, and $m_{t-1}$ contains the distilled memory returned from the previous step. 
To keep the reasoning context bounded, ActiveMem retains only the most recent $K$ interaction steps:
\begin{equation}
h_t = \mathrm{Trim}\left(h_{t-1} \cup \{a_t, o_t\}\right),
\label{eq:trimmed_history}
\end{equation}
where $a_t$ denotes the Planner's action and $o_t$ denotes the tool observation. $\mathrm{Trim}$ retains only the most recent $K$ interaction steps; the discarded content is not lost, as its gist has already been stored in the Memory Shards. The Planner maps this compact state to a set of retrieval queries, \textit{i.e.},

\begin{equation}
\mathcal{Q}_t = \pi(s_t),
\end{equation}
where each element $(q_i, k_i) \in \mathcal{Q}_t$ specifies a retrieval query $q_i$ and its retrieval budget $k_i$, \textit{i.e.}, the number of documents requested by the Planner for that query. After receiving the distilled gists from the memory system, the Planner updates its reasoning state and either issues another retrieval request or produces the final answer.

\subsection{Distributed Memory System}

The distributed memory system is designed to preserve document-level memory without overloading the Planner's reasoning context. It contains three components: Memory Shards that store persistent gists, parallel Memorizers that distill query-conditioned gists from retrieved documents, and an Operator that coordinates routing, reuse, and shard updates. Together, they provide a persistent and parallel memory layer for maintaining and reusing distilled information across the task.

\subsubsection{Memory Shards}

Memory Shards serve as the persistent storage layer of ActiveMem. Instead of merging information from different documents into a single compressed context, ActiveMem maintains a fixed set of logical shards throughout the task. Each shard $B_j$ is implemented as a key--value store indexed by document $c$. For each document entry, the shard stores the distilled gist $g_c$ and the set of queries that have previously used this document:
\begin{equation}
B_j[c] = \bigl(g_c,\; \mathcal{H}_c\bigr),
\quad
\mathcal{H}_c = \{q^{(1)}, q^{(2)}, \ldots\}
\label{eq:memory_shard}
\end{equation}

Because Memory Shards persist throughout the task and are logically isolated, distilled gists produced at any step remain available for later reasoning and are further enriched through asynchronous consolidation rather than discarded. This ensures that distilled information remains recoverable throughout the task while keeping the Planner's core reasoning context clean and compact, mitigating the trade-off between context overload and information loss that centralized agents face.

\subsubsection{Memorizers}

Inspired by the functional role of the hippocampus, each Memorizer is designed as a query-conditioned memory module. Given a query $q$ and the associated raw content $c$ routed by the Operator, the Memorizer produces a content-specific gist:
\begin{equation}
g_c = \omega(c,q), \quad g_c \in \mathcal{G} \cup \{\varnothing\},
\end{equation}
where $g_c \in \mathcal{G}$ denotes a valid gist when $c$ is relevant to $q$, and $g_c=\varnothing$ denotes an empty gist otherwise. The query $q$ provides top-down guidance by specifying which aspect of $c$ should be retained. The Memorizer filters out irrelevant information and returns only the distilled gist $g_c$ to the Planner. This prevents the Planner from directly processing raw content while preserving the information needed for the current reasoning step. Multiple Memorizers operate in parallel and write their outputs independently to their assigned Memory Shards.

\paragraph{Automatic Training Data Construction.}
A vanilla small language model often produces verbose responses and unnecessary reasoning traces when directly used as a Memorizer. We therefore construct supervised data from real agent interactions to train the Memorizer for concise and content-grounded distillation.

We follow the train/test split of BrowseComp-Plus introduced by Context-Folding~\citep{sun2025scaling}. On the training split, we run the full ActiveMem pipeline and collect 12{,}000 query-document pairs, denoted as $(q,c)$. Each pair contains a query $q$ issued by the Planner during task execution and a retrieved document $c$ routed to the memory system. These pairs are collected from actual agent trajectories, and therefore reflect the agent's genuine information needs during long-horizon reasoning.

For each query-document pair $(q,c)$, we use \texttt{gpt-oss-120b} as the teacher model to generate a concise semantic gist $g_c$ conditioned on the query, following the knowledge distillation paradigm~\citep{DBLP:journals/corr/HintonVD15}. The resulting triples $(q,c,g_c)$ are used as supervised training examples for the Memorizer.  We maintain a document-level separation between the training and test sets. Specifically, documents used to construct Memorizer training examples are excluded from the test-time document pool. This ensures that the Memorizer is evaluated on documents that do not overlap with its training corpus.

% To prevent data leakage, we enforce a strict document-level partition between the training and test sets. At test time, Memorizer-4B is not allowed to access any document that appears in the training corpus. This ensures that no test evidence is seen during training, either directly or through document overlap.

\paragraph{Memorizer Training.}
Given the supervised triples $(q,c,g_c)$, we fine-tune the Memorizer with a conditional next-token prediction objective:

\vspace{-5pt}
{\fontsize{8.8pt}{9.5pt}\selectfont
\begin{equation}
\mathcal{L}_{\mathrm{SFT}}
=
-\mathbb{E}_{(q,c,g_c)\sim\mathcal{D}}
\!\left[
\sum_t
\log p_{\omega}\!\left(g_{c,t}\mid q,c,g_{c,<t}\right)
\right],
\label{eq:memorizer_sft}
\end{equation}

}
where $\mathcal{D}$ denotes the Memorizer training set, $\omega$ denotes the trainable parameters of the Memorizer, and $g_{c,t}$ is the $t$-th token of the target gist $g_c$. This objective teaches the Memorizer to generate concise gists that are grounded in the retrieved document and conditioned on the Planner's query.

\subsubsection{Operator}

The Operator is the control layer between the Planner and the distributed memory system. It converts Planner queries and retrieved documents into memory tasks, routes them to the appropriate Memorizers, and manages when existing shard entries can be reused or updated. This allows ActiveMem to avoid redundant memory computation while keeping document-level memory up to date.

\paragraph{Memory Reuse.}
When a memory task $(q,c)$ arrives, the Operator first judges whether document $c$ has already been used to answer a similar query. It compares $q$ with the query history $\mathcal{H}_c$ stored in the corresponding shard:
\[
J(q,\mathcal{H}_c) \in \{\texttt{SIMILAR}, \texttt{NEW}\}.
\]
If the query is labeled as \texttt{SIMILAR}, the stored gist $g_c$ is returned directly to the Planner, avoiding an additional Memorizer call. If it is labeled as \texttt{NEW}, the task is dispatched to the Memorizer to produce a new query-conditioned gist.

\paragraph{Asynchronous Consolidation.}
The Operator additionally runs a consolidation pass that is decoupled from the main reasoning loop and does not block ongoing inference. When the same raw content has been distilled from multiple retrieval angles, the Operator merges the resulting gists into a single enriched shard entry that preserves all entities while eliminating redundancy:

\vspace{-12pt}
\begin{equation}
(g_c', \mathcal{H}_c') = \text{Consolidate}(g_c, g_{\text{new}}, \mathcal{H}_c, q),
\vspace{-12pt}
\end{equation}

\begin{algorithm}[!t]
\caption{ActiveMem Inference}
\label{alg:activemem}
\begin{algorithmic}[1]
\small
\renewcommand{\algorithmicrequire}{\textbf{Input:}}
\renewcommand{\algorithmicensure}{\textbf{Output:}}
\Require Question $x$, Max steps $T$, Shards $\{B_j\}$, Shard number $N$, History window $K$
\Ensure Answer $a$
\State \textit{Initialize all memory shards and set conversation history to system prompt and input question}
\For{$t = 1,\ldots,T$}
  \State $\mathcal{Q}_t \leftarrow \pi(s_t)$, \quad $s_t = (x, h_t, m_{t-1})$
  \If{$\mathcal{Q}_t = \texttt{submit\_answer}(a)$}
    \State \Return $a$
  \EndIf
  \State $D_t \leftarrow \textsc{Retrieve}(\mathcal{Q}_t)$
  \State $\mathcal{G}_t \leftarrow \emptyset$
  \For{\textbf{parallel} $(q, c, B_j) \in \textsc{Route}(D_t)$, at most $N$ concurrent}
      \If{$c \notin B_j$}
        \State \textit{Memory miss: distill $g \leftarrow \omega(c,q)$ and write $(g, q)$ to shard $B_j$}
      \ElsIf{$J(q, B_j[c].\mathcal{H}) = \texttt{SIMILAR}$}
        \State \textit{Memory hit: reuse $g \leftarrow B_j[c].g$ directly}
      \Else
        \State \textit{Memory hit with new query: distill $g \leftarrow \omega(c,q)$ and merge into $B_j$ asynchronously}
      \EndIf
      \State $\mathcal{G}_t[c] \leftarrow g$
  \EndFor
  \State $o_t \leftarrow \text{Serialize}(\mathcal{G}_t)$
  \State $h \leftarrow \text{Trim}(h \mathrel{+\!\!=} (m_t, o_t), K)$
\EndFor
\end{algorithmic}
\end{algorithm}

\subsection{ActiveMem Inference}

Algorithm~\ref{alg:activemem} summarizes the complete inference procedure of ActiveMem. At each step $t$, the Planner observes state $s_t = (x, h_t, m_{t-1})$---comprising the original question, the trimmed interaction history, and distilled memory returned from the previous step---and emits a set of retrieval queries $\mathcal{Q}_t$. If the Planner instead calls $\texttt{submit\_answer}(a)$, inference terminates and the answer $a$ is returned.

Otherwise, the Retriever fetches a document set $D_t$ in response to $\mathcal{Q}_t$, and $\textsc{Route}$ assigns each retrieved document to a stable Memory Shard. The assigned documents are processed by Memorizers in parallel,  with at most $N$ Memorizers running concurrently to match the number of available shards. For each document, the Operator applies one of three policies: a memory miss triggers fresh distillation and writes the new gist $(g, q)$ to the shard; a memory hit with a semantically similar query reuses the stored gist without invoking the Memorizer; a memory hit with a new query distills an updated gist and launches an asynchronous merge that consolidates old and new gists in the background without blocking the Planner.  After all documents are processed, the collected gists $\mathcal{G}_t$ are assembled into observation $o_t$, appended to history, and $\textsc{Trim}$ retains only the most recent $K$ interaction steps to keep the Planner context compact.

\section{Experiments}
\label{sec:experiments}

\subsection{Experimental Setup}

\vpara{Datasets.} We evaluate ActiveMem on two widely used benchmarks. \textbf{BrowseComp-Plus}~\citep{wei2025browsecomp} is our primary benchmark. We evaluate on 150 examples following the Easy, Medium, and Hard split introduced by Context-Folding~\citep{sun2025scaling}, which supports fine-grained analysis.
\textbf{GAIA}~\citep{mialon2024gaia} is a general-purpose agent benchmark. We use its WebSearch validation subset (106 examples) to focus the comparison on information-intensive retrieval tasks.

\vpara{Baselines.} We compare ActiveMem with two categories of baselines. \textit{Vanilla ReAct LLMs} include Kimi-K2.5 \citep{team2026kimi}, Qwen3.5-397B-A17B \citep{qwen3.5}, GLM-5.1 \citep{glm5team2026glm5vibecodingagentic}, and DeepSeek-V3.2 \citep{deepseek2025deepseek}, all operating under a standard ReAct loop without memory compression. To control context length, all vanilla baselines adopt a two-stage page-reading mechanism: the model first reads a 512-token preview; if deemed relevant, it issues an \texttt{open} action to load the full 4,096-token content. \textit{Centralized Memory Agents} include Context-Folding\citep{sun2025scaling}, AgentFold \citep{ye2025agentfold}, MemoBrain \citep{zhu2025memobrain}, and MemAgent\footnote{To align MemAgent with our task setting, we implement a sequential variant as a baseline to our parallel Memorizers by processing retrieved documents through MemAgent-7B.}\citep{yu2025memagent}. For a fair comparison, all models adopt Qwen3.5-397B-A17B as the backbone LLM, with all experiments executed under this unified setup.

\vpara{Metrics.} We evaluate the performance of the model using three metrics. \textbf{LasJ} (LLM-as-a-Judge) measures answer correctness using Qwen3.5-397B-A17B as the judge model.
\textbf{PFLOPs} measures total inference-time computational cost and serves as a model-agnostic proxy for both model scale and processed context length. Although token count is a widely used metric for measuring model efficiency, it masks crucial parameter-scale discrepancies across heterogeneous systems, making cross-model efficiency comparisons highly misleading. To address this, we strictly count linear FLOPs from matrix multiplications and quadratic FLOPs from self-attention, with architecture-specific adjustments for hybrid-attention and Dynamic Sparse Attention (DSA) models. 
% full derivations are given in Appendix~\ref{appendix:model_arch}. 
We also report total token count (in millions) alongside PFLOPs to provide a complementary view of context volume processed during inference.
\textbf{ACT} (Accuracy-Cost Trade-off) is a composite metric that balances model performance against computational cost. By penalizing raw accuracy with log-normalized PFLOPs, ACT effectively differentiates competitive methods based on their underlying efficiency. More details are provided in Appendix~\ref{appendix:metrics}.

\begin{table*}[t]
\centering
\small
\begin{tabular}{l cc cc cc ccc}
\toprule
\multirow{2}{*}{\textbf{Model}}
  & \multicolumn{2}{c}{\textbf{Easy}}
  & \multicolumn{2}{c}{\textbf{Medium}}
  & \multicolumn{2}{c}{\textbf{Hard}}
  & \multicolumn{3}{c}{\textbf{Total}}\\
\cmidrule(lr){2-3}\cmidrule(lr){4-5}\cmidrule(lr){6-7}\cmidrule(lr){8-10}
& LasJ & PFLOPs & LasJ & PFLOPs & LasJ & PFLOPs & LasJ & PFLOPs (tokens) & ACT\\
\midrule
\multicolumn{10}{l}{\textit{Vanilla ReAct LLMs}} \\
Kimi-k2.5            & \textbf{1.00} & 1023      & 0.80 & 6803      & 0.26 & 21643     & 0.69 & 29469 {\scriptsize (157M)}     & 0.640 \\
DeepSeek-V3.2        & \underline{0.98} & 1515     & 0.76 & 5078      & 0.34 & 8128      & 0.69 & 14721 {\scriptsize (140M)}     & 0.652 \\
GLM-5.1              & 0.96 & 1179      & 0.74 & 5213      & 0.34 & 10508     & 0.68 & 16901 {\scriptsize (158M)}     & 0.640 \\
Qwen3.5-397B-A17B    & \underline{0.98} & 1300     & 0.78 & 4999      & 0.22 & 12207     & 0.66 & 18506 {\scriptsize (297M)}     & 0.618 \\
\midrule
\multicolumn{10}{l}{\textit{Centralized Memory Agents}} \\
MemAgent             & 0.96 & 316       & 0.56 & 1507      & 0.26 & 2491      & 0.59 & 4314 {\scriptsize (167M)}      & 0.573 \\
MemoBrain            & \underline{0.98} & \underline{256}       & 0.70 & \textbf{464}       & 0.20 & \textbf{920}       & 0.63 & \textbf{1640} {\scriptsize (47M)}      & 0.630 \\
AgentFold            & 0.96 & 458       & 0.80 & 1466      & 0.30 & 3832      & 0.69 & 5757 {\scriptsize (116M)}      & 0.668 \\
Context-Folding      & 0.96 & 283       & \underline{0.84} & 1025      & \underline{0.36} & 2613      & \underline{0.72} & 3920 {\scriptsize (76M)}       & \underline{0.705} \\
\midrule
% \multicolumn{10}{l}{\textit{Ours}} \\
\textbf{ActiveMem (Ours)} & \textbf{1.00} & \textbf{204}       & \textbf{0.94} & \underline{653}      & \textbf{0.44} & \underline{1355}     & \textbf{0.79} & \underline{2145} {\scriptsize (188M)}     & \textbf{0.785} \\
\bottomrule
\end{tabular}
\caption{Overall evaluation on BrowseComp-Plus.
% \textbf{LasJ}: LLM-as-a-Judge accuracy; \textbf{PFLOPs}: estimated total inference-time compute; \textbf{ACT}: Accuracy-Cost Trade-off composite score (higher is better).
\textbf{Bold} and \underline{underline} indicate the best and second-best results.}
\label{tab:main_results}
\end{table*}

\begin{table}[t]
\centering
\small
\setlength{\tabcolsep}{4pt}
\resizebox{\columnwidth}{!}{%
\begin{tabular}{lccc}
\toprule
\textbf{Model} & \textbf{LasJ} & \textbf{PFLOPs (tokens)} & \textbf{ACT} \\
\midrule
\multicolumn{4}{l}{\textit{Vanilla ReAct LLMs}} \\
Kimi-k2.5         & 0.54 & 2442 {\scriptsize (32M)} & 0.497 \\
GLM-5.1           & 0.58 & 2233 {\scriptsize (28M)} & 0.539 \\
DeepSeek-V3.2     & 0.55 & 3839 {\scriptsize (47M)} & 0.500 \\
Qwen3.5-397B-A17B & 0.53 & 1007 {\scriptsize (15M)} & 0.502 \\
\midrule
\multicolumn{4}{l}{\textit{Centralized Memory Agents}} \\
AgentFold         & 0.59 & 535 {\scriptsize (15M)} & 0.573 \\
MemoBrain         & \underline{0.59} & 516 {\scriptsize (16M)} & \underline{0.573} \\
MemAgent          & 0.53 & \underline{363} {\scriptsize (13M)} & 0.520 \\
\midrule
% \multicolumn{4}{l}{\textit{Ours}} \\
\textbf{ActiveMem (Ours)}  & \textbf{0.62} & \textbf{187} {\scriptsize (12M)} & \textbf{0.620} \\
\bottomrule
\end{tabular}
}
\caption{Overall evaluation on GAIA.
% \textbf{LasJ}: LLM-as-a-Judge accuracy; \textbf{PFLOPs}: estimated total inference-time compute; \textbf{ACT}: Accuracy-Cost Trade-off composite score.
\textbf{Bold} denotes the best result, and \underline{underlining} denotes the second best.}
\label{tab:gaia_results}
\vspace{-12pt}
\end{table}

\subsection{Main Results}
As shown in Table~\ref{tab:main_results} and Table~\ref{tab:gaia_results}, we conclude three key observations: \textbf{(1) ActiveMem achieves the strongest accuracy--cost trade-off across both benchmarks.} 
On BrowseComp-Plus, ActiveMem obtains the highest overall LasJ score of 0.79 and the highest ACT score of 0.785. On GAIA, it also achieves the best LasJ score of 0.62 with the lowest computational cost.
The baseline results reveal two different limitations. Vanilla ReAct LLMs accumulate retrieved documents in the reasoning context, increasing context length and PFLOPs as inference proceeds. Centralized memory agents reduce context growth through compression, but may discard useful information from the document. MemoBrain illustrates this trade-off: on BrowseComp-Plus, its relatively low cost of 1,640 PFLOPs comes with a lower LasJ score of 0.63, while ActiveMem improves accuracy by +0.16 (0.63 $\rightarrow$ 0.79). On GAIA, ActiveMem matches MemoBrain's accuracy while reducing computation by 63.7\% (516 $\rightarrow$ 187 PFLOPs).
These results suggest that ActiveMem improves efficiency without sacrificing accuracy. By distilling retrieved documents in parallel and preserving reusable document-specific gists in Memory Shards, ActiveMem keeps the Planner's context compact while retaining information needed for later reasoning.
\textbf{(2) ActiveMem's advantage is more pronounced on the BrowseComp-Plus Hard split.}
On Easy questions, most methods are close to saturation, leaving limited room for differentiation. As the difficulty level increases, ActiveMem shows a clearer advantage. On the Medium split, ActiveMem reaches 0.94, outperforming the best baseline by +0.10 absolute LasJ score (0.84 $\rightarrow$ 0.94), corresponding to a relative improvement of +11.9\%. On the Hard split, ActiveMem achieves 0.44, improving over the best baseline by +0.08 absolute LasJ score (0.36 $\rightarrow$ 0.44), corresponding to a relative improvement of +22.2\%.
This result highlights the benefit of decoupled memory under more difficult retrieval-intensive reasoning. By returning distilled gists instead of raw retrieved documents, ActiveMem keeps the Planner's context compact and reduces the degradation caused by long contexts or lossy centralized compression.
\begin{table}[!t]
\centering
\small
\begin{tabular}{lccc}
\toprule
\textbf{Module} & \textbf{Tokens (M)} & \textbf{PFLOPs} & \textbf{Share} \\
\midrule
Planner   & 16.69  & 606   & 28.3\% \\
Memorizer & 163.88 & 1478  & 68.9\% \\
Operator  & 7.44   & 61    & 2.8\%  \\
\midrule
Total     & 188.14 & 2145  & 100\%  \\
\bottomrule
\end{tabular}
\caption{Module-level breakdown of token usage and PFLOPs for ActiveMem on BrowseComp-Plus.}
\label{tab:cost_breakdown}
\vspace{-13pt}
\end{table}
\textbf{(3) Offloading token-intensive processing to lightweight Memorizers enables more favorable test-time scaling at lower cost.}
As shown in Table~\ref{tab:cost_breakdown}, Memorizers account for 68.9\% of total PFLOPs while the Planner contributes only 28.3\%. By routing most token processing to small models, ActiveMem processes substantially more tokens than centralized agents (188M vs.\ 47--167M on BrowseComp-Plus) while keeping total PFLOPs low, since the marginal cost per token is much lower for a 4B Memorizer than for the large Planner. The higher token consumption on BrowseComp-Plus reflects the benchmark's tighter question constraints, which demand broader evidence gathering; ActiveMem's ability to process more evidence at lower cost is consistent with the test-time scaling principle~\citep{DBLP:journals/corr/abs-2408-03314}. On GAIA, where questions are less constrained and agents converge faster, token counts are uniformly lower across all methods. Case studies are provided in Appendix~\ref{appendix:additional_analysis}.
Figure~\ref{fig:pflops_vs_steps} shows that ActiveMem scales better as the number of reasoning steps increases. Its per-case cost curve remains flatter than those of the baselines, and the cost gap widens at higher step counts, reflecting lower sensitivity to context length. Context-Folding is an exception, as its lower cost stems from early branch termination rather than reduced per-step computation.

\begin{figure}[t]
\centering
\includegraphics[width=\columnwidth]{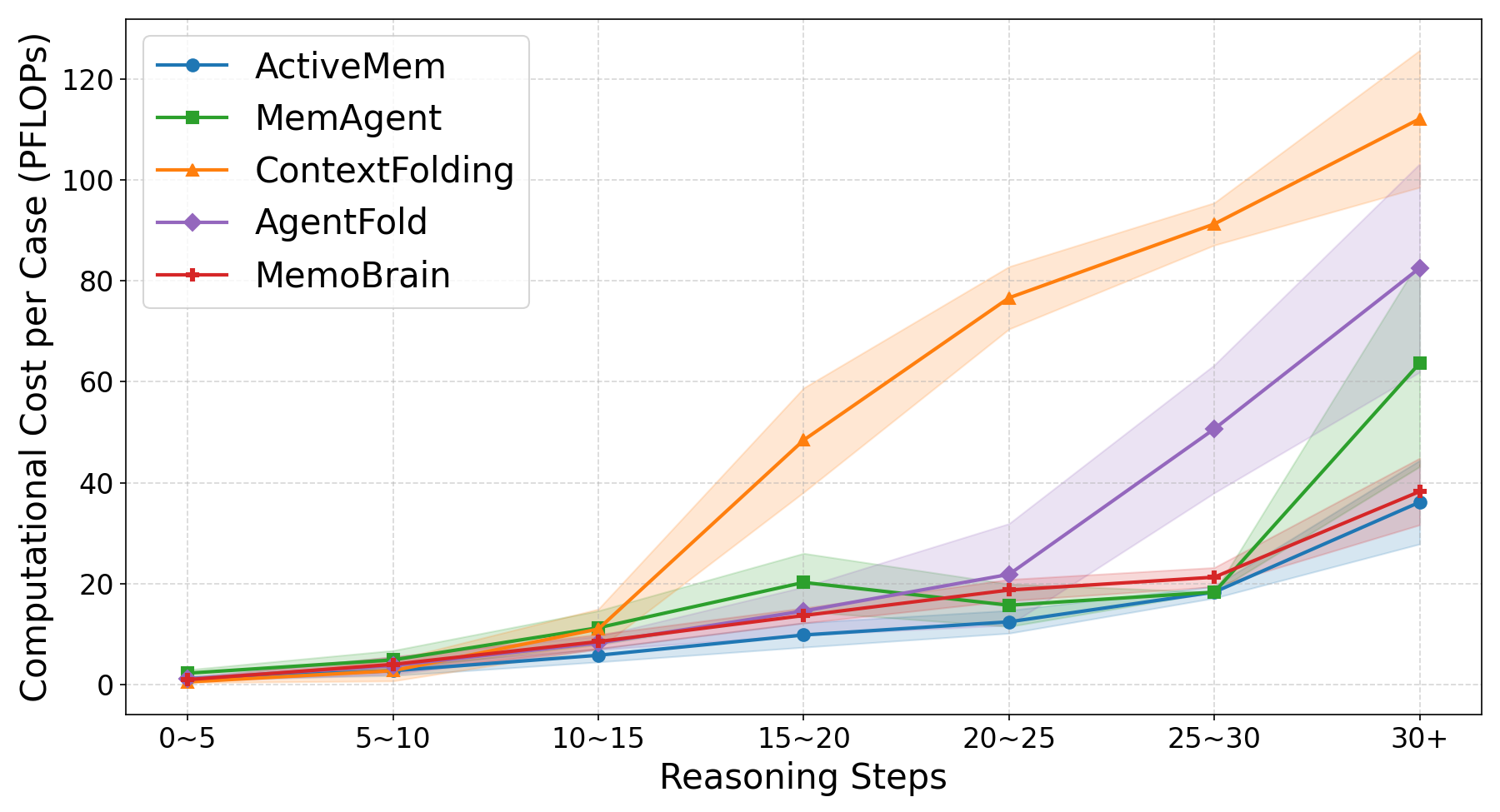}
\caption{Average computational cost per case (PFLOPs) across reasoning step ranges. Shaded regions indicate $\pm$1 standard deviation.}
\label{fig:pflops_vs_steps}
\end{figure}

\subsection{Ablation Studies}

We ablate two main design choices in ActiveMem: persistent Memory Shards and Memorizer variants. Additional analyses of Trim window size, Memorizer scale trade-offs, and Memory Consolidation are in Appendix~\ref{appendix:additional_analysis}.

\subsubsection{Memory Shards Reduce Redundancy and Preserve Information}

We ablate Memory Shards by removing shard-based storage and retrieval while keeping the Memorizer unchanged. We define this variant as \textit{ActiveMem w/o Shards}, where distilled gists are returned directly to the Planner at each step rather than stored for selective reuse. The comparative results are reported in Table~\ref{tab:ablation_memory}.

\hide{To isolate the contribution of the persistent memory structure, we ablate it by removing all memory storage and retrieval: the Memorizer still distills semantic gists from raw content, but these gists are returned directly to the Planner in full at the end of each step rather than being stored and selectively retrieved across steps. This variant, denoted ActiveMem w/o PM, preserves the distillation capability while eliminating the structured memory layer entirely. Results are shown in Table~\ref{tab:ablation_memory}.}

\begin{table}[t]
\centering
\setlength{\tabcolsep}{4pt}
\resizebox{\columnwidth}{!}{%
\begin{tabular}{lcccc}
\toprule
\textbf{Model} & \textbf{P. PFLOPs} & \textbf{M. PFLOPs} & \textbf{Tot. PFLOPs} & \textbf{LasJ} \\
\midrule
ActiveMem w/o Shards & \textbf{546} & 1739 & 2285 & 0.750 \\
ActiveMem            & 606 & \textbf{1478} & \textbf{2145} & \textbf{0.786} \\
\bottomrule
\end{tabular}
}
\caption{Ablation on Memory Shards. P., M., and Tot. denote Planner, Memorizer, and total inference cost (PFLOPs), respectively.}
\label{tab:ablation_memory}
\vspace{-12pt}
\end{table}

\hide{Removing persistence leads to a drop in both accuracy and efficiency. Without a persistent memory store, when the Planner later issues a query overlapping with previously processed content, the Operator cannot retrieve a stored gist and must instead re-dispatch the raw content to the Memorizer for re-distillation, driving up total PFLOPs. On the accuracy side, the absence of persistent memory means that gists derived from earlier reasoning steps cannot be reliably carried forward, forcing the Planner to reason with an incomplete information state. These results demonstrate that distillation and persistence are complementary: distillation compresses raw content into reusable semantic gists, while persistent memory ensures those gists are efficiently retrieved without redundant computation.}

Removing Memory Shards reduces both accuracy and efficiency. Without shard-based storage, overlapping queries trigger redundant Memorizer passes on the same raw content, raising Memorizer PFLOPs from 1478 to 1739. At the same time, previously distilled gists are not consolidated back into the Planner's context on new queries---keeping Planner context smaller (546 vs.\ 606 PFLOPs) but causing information loss that degrades final accuracy.
These results demonstrate the complementary roles of distillation and Memory Shards: the former compresses content into semantic gists, while the latter persists and reuses these gists across reasoning steps, eliminating redundant computation. Together, they also preserve information across reasoning steps, contributing to higher accuracy.

\subsubsection{A Stronger Memorizer Improves Reasoning Quality}

We compare three 4B-scale Memorizer variants with all other components fixed: Qwen3-4B-Instruct-2507 (instruction-tuned), Qwen3-4B~\citep{DBLP:journals/corr/abs-2505-09388} (vanilla reasoning), and our SFT-trained Memorizer-4B. 

As shown in Table~\ref{tab:ablation_memorizer}, the choice of Memorizer design is critical to final task performance. Compared with the Qwen3-4B-Instruct-2507, both reasoning-oriented variants achieve higher LasJ accuracy. This suggests that a reasoning-capable Memorizer can produce more precise and compact gists for the Planner. The effect is also reflected in Planner cost: the two reasoning-oriented variants substantially reduce Planner PFLOPs, indicating that a cleaner and less redundant distilled context makes downstream reasoning more efficient. This improvement comes with a higher Memorizer cost, since reasoning-oriented Memorizers spend more computation on their own intermediate reasoning. However, the additional cost is offset by consistent gains in answer accuracy. SFT further improves this trade-off. Compared with vanilla Qwen3-4B, Memorizer-4B makes distillation more concise and task-aligned, siginificantlyreducing Memorizer cost from 1833 to 1478 PFLOPs and Planner cost to 606 PFLOPs, while improving LasJ to 0.786.

\begin{table}[!t]
\centering
\setlength{\tabcolsep}{4pt}
\resizebox{\columnwidth}{!}{%
\begin{tabular}{lcccc}
\toprule
\textbf{Type} & \textbf{P. PFLOPs} & \textbf{M. PFLOPs} & \textbf{Tot. PFLOPs} & \textbf{LasJ} \\
\midrule
Instruct & 1120 & \textbf{1378} & 2659 & 0.667 \\
Thinking (Vanilla) & 731 & 1833 & 2647 & 0.733 \\
Thinking (SFT) & \textbf{606} & 1478 & \textbf{2145} & \textbf{0.786} \\
\bottomrule
\end{tabular}
}
\caption{Ablation on Memorizer variants.}
\label{tab:ablation_memorizer}
\vspace{-12pt}
\end{table}
\section{Conclusion}
\label{sec:conclusion}
We introduce ActiveMem, a neuroscience-inspired framework that decouples memory management from high-level reasoning. ActiveMem separates executive reasoning from parallel active memory consolidation, alleviating the trade-off between context overload and information loss in centralized memory designs. Extensive evaluations on diverse benchmarks show that ActiveMem achieves higher accuracy than competitive baselines while substantially reducing computational overhead. These results validate our distributed design and suggest that decoupled active memory provides an efficient and scalable foundation for long-horizon reasoning.

\clearpage
\section*{Limitations}

There are two limitations to this work. First, although we report computational efficiency metrics, we do not directly measure deployment-level efficiency, such as wall-clock latency or monetary cost. These quantities are highly sensitive to serving infrastructure, API response variability, rate limits, hardware utilization, and provider-specific pricing assumptions. Since our experiments involve both commercial API models and locally deployed open-source models, a fair comparison would require a fully controlled deployment setting with consistent serving infrastructure. We leave such profiling to future work.
Second, the generalizability of ActiveMem beyond the evaluated domains remains an open question. The Memorizer is trained via supervised fine-tuning on agent interaction data collected from BrowseComp-Plus, a web-search benchmark consisting mainly of tightly constrained factual questions. While this leads to effective memory distillation on the evaluated tasks, its applicability to other distinct paradigms, such as mathematical reasoning, code generation, or multimodal agent workflows, has not been systematically cross-validated. Adapting ActiveMem to new domains may require collecting domain-specific interaction data and further tuning the Memorizer.

% \clearpage
\bibliography{custom}
\clearpage
\appendix
% ─────────────────────────────────────────────
\section{Appendix}
\label{appendix:impl}
% ─────────────────────────────────────────────

\subsection{Implementation Details}
\label{appendix:impl_details}

ActiveMem instantiates the Planner with Qwen3.5-397B-A17B, the Memorizer with Memorizer-4B, the Operator with Qwen3-4B, and the retrieval module with Qwen3-Embedding-8B.
Table~\ref{tab:impl_details} shows the model and parameter settings for ActiveMem and all four baselines. To ensure a fair comparison, all systems share the same Planner backbone (Qwen3.5-397B-A17B) with identical temperature and step budget. For each baseline, hyperparameters not listed in the table follow the settings reported in the respective original paper.

All SFT training was conducted on 4 NVIDIA A100 (80GB) GPUs. For inference, the Planner model (Qwen3.5-397B-A17B) was accessed via API due to its large parameter count, while all other models---including the Memorizer, Operator, and retrieval module---were each served locally on a single NVIDIA A100 (80GB) GPU.
All models and datasets used in this work are publicly available and used in accordance with their respective licenses for non-commercial research purposes.

\subsection{Prompts}
\label{appendix:prompts}

We provide the full prompts for all four model roles in ActiveMem. The \textit{Planner} prompt defines a three-step decision flow governing when to search versus when to submit a final answer. The \textit{Operator: Similarity Judge} prompt instructs a lightweight model to detect query overlap so that stored gists can be reused without redundant retrieval. The \textit{Operator: Memory Consolidation} prompt merges an existing document gist with a newly produced one while preserving all factual content. The \textit{Memorizer} prompt directs the extraction model to produce a concise, document-grounded gist for a given sub-query, outputting \texttt{NONE} only when no relevant information is present.

\tcbset{
  promptbox/.style={
    breakable, enhanced,
    colframe=black!50,
    colback=black!4!white,
    colbacktitle=black!50,
    coltitle=white,
    boxrule=0.8pt,
    arc=2pt,
    left=6pt, right=6pt, top=4pt, bottom=4pt,
    fontupper=\footnotesize,
    fonttitle=\bfseries\small,
    width=\columnwidth,
  }
}

\begin{tcolorbox}[promptbox, title={Planner}]
You are a strategic task planner. Your job is to decide what to search
for in the corpus or when to answer, following a clear three-step decision flow.

\medskip\noindent\textbf{\#\# Tools (function calling)}\\[2pt]
\textbullet\ \texttt{advance\_search(tasks)}: parallel dense retrieval + worker analysis.\\
\quad\textbullet\ Input: \texttt{\{"tasks": [\{"query": string, "topk": int\}, ...]\}}\\
\quad\textbullet\ 1--4 tasks per call; each task must target a DIFFERENT aspect or
  candidate (different person, relation, time/place, etc.).\\
\quad\textbullet\ \texttt{3 <= topk <= 8}; use larger topk only when necessary.\\[2pt]
\textbullet\ \texttt{submit\_answer(answer)}: when you already have enough evidence,
  submit a short English phrase that directly answers the user's question.

\medskip\noindent\textbf{\#\# Decision flow (follow in order)}\\[2pt]
\textbf{Step 1 -- Decompose the question:}\\
\textbullet\ Break down the user question into the minimal set of information pieces
  needed to answer it (who\,/\,what\,/\,when\,/\,where\,/\,relations, etc.).\\
\textbullet\ Think of these as independent sub-questions or search paths.

\medskip\noindent\textbf{Step 2 -- Match against what you already know:}\\
\textbullet\ Use the latest tool results (provided as messages from tools) to decide
  which pieces are already satisfied and which are still missing.\\
\textbullet\ If current knowledge is enough to form a complete, precise answer,
  call \texttt{submit\_answer} with a concise, direct answer phrase.
  Do NOT search again in that case.

\medskip\noindent\textbf{Step 3 -- Organize search queries for parallel retrieval:}\\
\textbullet\ For any missing pieces, prepare search tasks for \texttt{advance\_search}.\\
\textbullet\ Each \texttt{task.query} must be a SHORT English keyword phrase, not a full
  sentence, combining key entities\,/\,dates\,/\,relations.\\
\textbullet\ Avoid near-duplicate queries; each task should explore a distinct
  hypothesis, angle, or candidate.

\medskip\noindent\textbf{Output rules:}\\
\textbullet\ NEVER fabricate tool outputs; always wait for real tool responses.\\
\textbullet\ NEVER answer in free-form prose; every final answer must be returned
  via \texttt{submit\_answer}.\\
\textbullet\ Keep internal reasoning minimal; your visible output is ONLY function
  calls chosen according to the flow above.
\end{tcolorbox}

\begin{tcolorbox}[promptbox, title={Operator: Similarity Judge}]
You are a strict semantic-equivalence judge for short information-seeking
queries over the SAME document. You will receive a NEW sub-query and a list
of PREVIOUS sub-queries that were already answered using this document.

\medskip\noindent
Decide whether the NEW sub-query is asking for essentially the same
information aspect(s) as ANY of the previous sub-queries, so that the
stored summary is already sufficient.

\medskip\noindent\textbf{Decision rules:}\\
\textbullet\ Output \texttt{SIMILAR} if the NEW sub-query targets the same entity(ies) AND the
  same attribute/relation/time-scope as at least one previous sub-query.
  Minor paraphrase, synonym, or word order changes $\to$ \texttt{SIMILAR}.\\
\textbullet\ Output \texttt{NEW} if the NEW sub-query targets a different attribute, a
  different time point, a different entity, or asks for strictly more
  specific information than any previous one.\\
\textbullet\ When in doubt, output \texttt{NEW}.

\medskip\noindent\textbf{Output format (STRICT):}\\
\textbullet\ Output ONLY one token: either exactly \texttt{SIMILAR} or exactly \texttt{NEW}.\\
\textbullet\ Do NOT include quotes, punctuation, reasoning, or any extra text.
\end{tcolorbox}

\begin{tcolorbox}[promptbox, title={Operator: Memory Consolidation}]
You are a memory consolidator. You will receive:\\
\textbullet\ an OLD summary previously produced for this document\\
\textbullet\ a NEW summary just produced for this document (for a different sub-query)\\
\textbullet\ the list of sub-queries that led to OLD\\
\textbullet\ the new sub-query that led to NEW

\medskip\noindent
Your job: produce ONE merged summary that preserves all document-grounded
facts from both OLD and NEW, de-duplicating identical statements.

\medskip\noindent\textbf{Hard rules:}\\
\textbullet\ Do NOT add any external knowledge or speculation.\\
\textbullet\ Do NOT drop factual content that is present in OLD or NEW.\\
\textbullet\ Prefer concrete entities, dates, and relations over vague wording.\\
\textbullet\ Keep the merged summary concise: at most 5 sentences, ${\le}$150 words.\\
\textbullet\ Output ONLY the merged summary text. No prefixes, no bullet points,
  no explanation, no quotes.
\end{tcolorbox}

\begin{tcolorbox}[promptbox, title={Memorizer}]
You are a precise information extraction assistant for a single document.
You will receive ONE sub-query phrase and ONE document snippet.
Your job is to write a SHORT summary of information in THIS document that is
relevant to the sub-query.

\medskip\noindent\textbf{VERY IMPORTANT:}\\
\textbullet\ If the document contains ANY information related to the sub-query
  (even partially), you MUST write a summary.\\
\textbullet\ You may ignore parts of the sub-query that are NOT supported by the document.\\
\textbullet\ Output `NONE' ONLY when the document contains NO information related
  to the sub-query at all.\\
\textbullet\ Do NOT guess or add external knowledge.

\medskip\noindent\textbf{Output rule:}\\
\textbullet\ If there is some relevant information: output ONLY a concise summary
  in 1--2 sentences (${\le}$40 words), no bullet points, no explanation.\\
\textbullet\ If there is truly no relevant information: output EXACTLY `NONE'.
\end{tcolorbox}
\begin{table*}[t]
\centering
\small
\setlength{\tabcolsep}{4pt}
\begin{tabular}{llll}
\toprule
\textbf{System} & \textbf{Component} & \textbf{Parameter} & \textbf{Value} \\
\midrule
\multirow{16}{*}{ActiveMem}
  & \multirow{4}{*}{Planner}
    & Model              & Qwen3.5-397B-A17B \\
  & & Temperature        & 0.3 \\
  & & Max steps $T$      & 50 \\
  & & History window $K$ & 10 \\
\cmidrule{2-4}
  & \multirow{3}{*}{Memorizer}
    & Model              & Memorizer-4B \\
  & & Temperature        & 0.2 \\
  & & Max doc.\ tokens   & 4{,}096 \\
\cmidrule{2-4}
  & \multirow{5}{*}{Operator}
    & Model              & Qwen3-4B \\
  & & Judge temperature  & 0.0 \\
  & & Judge max tokens   & 8 \\
  & & Consol.\ temp.     & 0.2 \\
  & & Consol.\ max tokens & 1{,}024 \\
\cmidrule{2-4}
  & Retrieval
    & Embedding model    & Qwen3-Embedding-8B \\
\cmidrule{2-4}
  & \multirow{2}{*}{Memory}
    & Memory shards $B$  & 16 \\
  & & Consol.\ pool size & 16 \\
\midrule
\multirow{5}{*}{AgentFold}
  & \multirow{3}{*}{Planner}
    & Model              & Qwen3.5-397B-A17B \\
  & & Temperature        & 0.3 \\
  & & Max steps $T$      & 50 \\
\cmidrule{2-4}
  & \multirow{2}{*}{Summary LLM}
    & Model              & gpt-oss-120b (${\sim}5.1$B active) \\
  & & Temperature        & 0.2 \\
\midrule
\multirow{3}{*}{Context-Folding}
  & \multirow{3}{*}{Planner}
    & Model              & Qwen3.5-397B-A17B \\
  & & Temperature        & 0.3 \\
  & & Max steps $T$      & 50 \\
\midrule
\multirow{7}{*}{MemoBrain}
  & \multirow{3}{*}{Planner}
    & Model              & Qwen3.5-397B-A17B \\
  & & Temperature        & 0.3 \\
  & & Max steps $T$      & 50 \\
\cmidrule{2-4}
  & \multirow{2}{*}{MemoBrain}
    & Model              & MemoBrain-8B \\
  & & Temperature        & 0.7 \\
\cmidrule{2-4}
  & \multirow{2}{*}{Auxiliary}
    & Model              & Qwen3-30B-A3B-Instruct-2507 \\
  & & Temperature        & 0.7 \\
\midrule
\multirow{5}{*}{MemAgent}
  & \multirow{3}{*}{Planner}
    & Model              & Qwen3.5-397B-A17B \\
  & & Temperature        & 0.3 \\
  & & Max steps $T$      & 50 \\
\cmidrule{2-4}
  & \multirow{2}{*}{MemAgent}
    & Model              & RL-MemAgent-7B \\
  & & Temperature        & 0.3 \\
\bottomrule
\end{tabular}
\caption{Implementation details for all evaluated systems.}
\label{tab:impl_details}
\end{table*}
% ─────────────────────────────────────────────
\subsection{Metrics}
\label{appendix:metrics}
% ─────────────────────────────────────────────

\subsubsection{PFLOPs Computation Details}
\label{appendix:model_arch}

\paragraph{Why PFLOPs.}
Several cost metrics are commonly used in agent evaluation, each with limitations.
\textit{Token count} is the most prevalent proxy and maps directly to API billing.
However, token count ignores model scale: a single token processed by Qwen3.5-397B-A17B incurs far more computation than one processed by a 4B model, making cross-system token comparisons misleading when the underlying models differ.
\textit{LLM call count} is even coarser---it conflates calls to models of vastly different sizes and cannot distinguish a lightweight coordination call from a full reasoning pass.
\textit{Wall-clock latency} is the most intuitive measure for deployment but is strongly hardware-dependent and difficult to reproduce across different infrastructure configurations, making it unsuitable for fair academic comparison.
PFLOPs addresses these shortcomings by jointly accounting for model scale (via $N_{\text{act}}$) and context length (via the quadratic attention term), yielding a hardware-agnostic, model-agnostic estimate of total computation that enables fair comparison across heterogeneous systems. We note that PFLOPs is itself an approximation: it counts only the two dominant arithmetic terms (linear projections and full attention matmuls) and does not capture memory-bandwidth costs, KV cache effects, or auxiliary operations. Nevertheless, as a relative ranking metric applied uniformly across all systems, it provides a principled and reproducible basis for comparison.

\paragraph{Formula.}
For a single LLM call with $S_{in}$ prefill tokens and $S_{out}$ decode tokens, let $T = S_{in}+S_{out}$. The FLOPs for that call are estimated as:
\[
\begin{aligned}
\text{FLOPs}
&= 2 N_{\text{act}} T \\
&\quad + L_{\text{full}} \cdot n_{\text{heads}}
\cdot (d_{QK}+d_V) \cdot T^2 ,
\end{aligned}
\]
and reported as \(\text{PFLOPs}=\text{FLOPs}/10^{15}\).
Total PFLOPs per case are obtained by summing this estimate over all LLM calls made during inference, across all modules (Planner, Memorizer, and Operator).
Here \(N_{\text{act}}\) denotes the number of parameters activated per token,
including dense attention/projection parameters and activated MoE/shared expert
parameters. For dense models, \(N_{\text{act}}\) equals the total parameter count.
The first term approximates the FLOPs of linear projections and FFN/MoE
matrix multiplications, while the second term captures the dominant quadratic
cost of causal full attention, with the causal mask reducing the effective attention window by half (canceling the factor of~2 from the matrix multiply).

For MLA models, we use \(d_{QK}=d_{\text{nope}}+d_{\text{rope}}\), yielding an
architecture-normalized estimate rather than an exact kernel-level FLOP count.
For hybrid-attention models, only the \(L_{\text{full}}\) full-attention layers
contribute to the quadratic term; linear-attention layers are treated as
sequence-linear costs and are absorbed into the linear-time approximation.

For models with Dynamic Sparse Attention (DSA), the attention term is replaced by:
\[
\begin{aligned}
\text{Attn}_{\text{DSA}}
&= C_{\text{idx}} \cdot T^2 \\
&\quad + C_{\text{main}} \cdot T \cdot \min(T,k_{\text{top}}),
\end{aligned}
\]
where \(C_{\text{idx}} = L \cdot n_{\text{idx}} \cdot d_{\text{idx}}\) denotes the
cost coefficient of the lightweight indexer,
\(C_{\text{main}} = L \cdot n_{\text{heads}} \cdot (d_{QK}+d_V)\) denotes the
cost coefficient of the main sparse attention over the selected candidates,
and \(k_{\text{top}}\) is the number of KV candidates selected by the indexer
(\(k_{\text{top}}=2048\) for both DeepSeek-V3.2 and GLM-5.1).\footnote{The formula counts only the two dominant FLOPs terms. Auxiliary operations (RMSNorm, SiLU activations, MoE routing, embedding lookup) are not separately itemized, as they are not expected to change the relative ranking across systems dominated by linear projections and attention matmuls. KV cache effects (e.g., reduced attention cost from prefix caching or paged attention) are likewise excluded: such behavior is highly infrastructure-dependent and difficult to estimate reliably across different deployment configurations. We apply the same formula uniformly to all systems to ensure a fair and reproducible comparison.}

\subsubsection{ACT: Accuracy--Cost Tradeoff}
\label{appendix:act}

A single accuracy metric cannot distinguish between a system that achieves high accuracy at modest cost and one that achieves the same accuracy by spending orders of magnitude more computation. Conversely, a pure efficiency metric favors systems that answer cheaply at the expense of correctness. ACT (Accuracy--Cost Tradeoff) is designed to capture both dimensions in a single scalar that reflects the accuracy--efficiency operating point of each system.

\paragraph{Cost normalization.}
Raw PFLOPs values span several orders of magnitude across systems (e.g., from ${\sim}200$ to ${\sim}30{,}000$ PFLOPs in our BrowseComp-Plus results). A linear normalization would give disproportionate weight to the few high-cost outliers. We therefore apply log-normalization:
{\small\[
\text{CostNorm}_i = \frac{\log(\text{PFLOPs}_i)-\min_j \log(\text{PFLOPs}_j)}{\max_j \log(\text{PFLOPs}_j)-\min_j \log(\text{PFLOPs}_j)},
\]}
which maps all systems to $[0,1]$ and treats equal multiplicative differences in cost as equal differences on the normalized scale---a natural choice since computational cost tends to scale multiplicatively with context length and model size.

\paragraph{ACT definition.}
ACT is defined as a penalized accuracy:
\[
\text{ACT}_i = \text{LasJ}_i - \alpha \cdot \text{CostNorm}_i,
\]
where $\alpha = 0.05$ is the penalty weight. LasJ accuracy is the primary objective, and a system that achieves the highest accuracy should still rank highly even at elevated cost. The role of the cost term is to break near-ties in accuracy in favor of the more efficient system, and to flag systems whose marginal accuracy gain does not justify the additional computation. At $\alpha=0.05$, the most expensive system (CostNorm${}=1$) incurs a maximum penalty of 5 percentage points, which is approximately the minimum meaningful accuracy gap observed in our benchmarks. CostNorm is computed separately within each benchmark so that ACT values are directly comparable across rows within a table but should not be compared across tables.

\paragraph{Interpretation.}
A system with higher ACT achieves a better accuracy--efficiency operating point: either higher accuracy at comparable cost, or comparable accuracy at substantially lower cost. Because normalization is benchmark-relative, adding or removing a system can shift all CostNorm values; ACT should therefore be interpreted in the context of the full comparison set rather than as an absolute score.

% ─────────────────────────────────────────────
\subsection{Additional Analysis}
\label{appendix:additional_analysis}
% ─────────────────────────────────────────────

\subsubsection{Trim Window Size Balances Planner Context and Efficiency}
\label{sec:ablation_trim}

The Trim mechanism controls how many recent interaction steps are retained in the Planner's context, with the window size denoted by $k$. To assess its effect, we vary $k \in \{5, 10, 15\}$ and compare these settings with a no-Trim variant, \textit{ActiveMem w/o Trim}, as shown in Table~\ref{tab:ablation_trim}.

\begin{table}[t]
\centering
\small
\setlength{\tabcolsep}{4pt}
\resizebox{\columnwidth}{!}{%
\begin{tabular}{lcccc}
\toprule
\textbf{Trim $k$} & \textbf{P. PFLOPs} & \textbf{M. PFLOPs} & \textbf{Tot. PFLOPs} & \textbf{LasJ} \\
\midrule
$k=5$        & \textbf{446.20}  & 1718.79 & 2253.84 & 0.740 \\
$k=10$       & 606.04  & \textbf{1477.56} & \textbf{2144.55} & \textbf{0.786} \\
$k=15$       & 833.13  & 1580.96 & 2493.78 & 0.733 \\
w/o Trim     & 1225.13 & 1519.63 & 2801.09 & 0.747 \\
\bottomrule
\end{tabular}
}
\caption{Ablation on Trim window size.}
\label{tab:ablation_trim}
\end{table}

We observe that $k=10$ provides the best accuracy--efficiency trade-off. When $k$ is too small (\textit{i.e.}, $k=5$), the Planner receives insufficient recent context and must compensate by issuing more retrieval queries, substantially increasing Memorizer load (1718.79 PFLOPs) and degrading accuracy. When $k$ is too large (\textit{i.e.}, $k=15$) or absent entirely, Planner cost grows sharply as it processes longer and noisier context windows, without a corresponding accuracy gain. These results confirm that $k=10$ is a well-calibrated operating point and that the Trim mechanism is robust within a moderate range around this value.

\subsubsection{Memorizer Scale Trades Accuracy Against Efficiency}
\label{appendix:memorizer_scale}

We vary the Memorizer across three parameter scales (0.6B, 1.7B, and 4B) while holding all other components fixed, in order to characterize how model size shapes the accuracy--efficiency trade-off. Table~\ref{tab:ablation_scale} reports the results on BrowseComp-Plus.

\begin{table}[h]
\centering
\small
\setlength{\tabcolsep}{4pt}
\resizebox{\columnwidth}{!}{%
\begin{tabular}{lcccc}
\toprule
\textbf{Scale} & \textbf{P. PFLOPs} & \textbf{M. PFLOPs} & \textbf{Tot. PFLOPs} & \textbf{LasJ} \\
\midrule
Memorizer-0.6B & 744 & \textbf{371} & \textbf{1201} & 0.560 \\
Memorizer-1.7B & 642 & 780 & 1498 & 0.687 \\
Memorizer-4B & \textbf{606} & 1478 & 2145 & \textbf{0.786} \\
\bottomrule
\end{tabular}
}
\caption{Ablation on Memorizer scale on BrowseComp-Plus (LasJ).}
\label{tab:ablation_scale}
\end{table}

The results reveal a non-trivial interaction between Memorizer scale and Planner efficiency. While reducing Memorizer scale substantially lowers Memorizer cost (from 1,478 to 371 PFLOPs for Memorizer-4B vs.\ 0.6B), Planner cost moves in the opposite direction, rising from 606 to 744 PFLOPs. This counter-intuitive pattern stems from a capability gap: a weaker 0.6B Memorizer fails to reliably capture critical information during distillation, so the gists it produces are incomplete or misleading. The Planner, receiving lower-quality evidence, is forced into more rounds of suboptimal reasoning---issuing additional retrieval queries, re-examining previously processed content, and converging more slowly---which drives up its own cost. The final LasJ of 0.560 confirms that this additional Planner effort fails to compensate for the information loss at the distillation stage.

Viewed across all three scales, the Planner cost decreases monotonically as Memorizer scale increases (744 $\to$ 642 $\to$ 606 PFLOPs), indicating that a stronger Memorizer produces more precise and complete gists, reducing the Planner's reasoning burden at every step. This suggests that investing compute in a more capable Memorizer yields compounding returns: lower Planner cost, higher accuracy, and a more predictable overall computation profile. Memorizer-0.6B reduces total PFLOPs by roughly 44\% relative to Memorizer-4B, but at the cost of a 22.6\% point accuracy drop---a disproportionate exchange that makes aggressive scale reduction an unfavorable operating point in practice.

\subsubsection{Memory Consolidation Keeps Planner Context Compact and Clean}
\label{appendix:ablation_conso}

To isolate the effect of the Operator's asynchronous consolidation mechanism, we evaluate \textit{ActiveMem (w/o Conso.)}, a variant in which newly produced gists are appended directly to the existing shard content without any merging or deduplication. Table~\ref{tab:ablation_conso} compares this variant against the full ActiveMem on BrowseComp-Plus.

\begin{table}[h]
\centering
\small
\setlength{\tabcolsep}{4pt}
\resizebox{\columnwidth}{!}{%
\begin{tabular}{lcccc}
\toprule
\textbf{System} & \textbf{P. PFLOPs} & \textbf{M. PFLOPs} & \textbf{Tot. PFLOPs} & \textbf{LasJ} \\
\midrule
ActiveMem (w/o Conso.) & 767.74          & 1543.48          & 2311.22          & 0.720 \\
ActiveMem             & \textbf{606.04} & \textbf{1477.56} & \textbf{2144.55} & \textbf{0.786} \\
\bottomrule
\end{tabular}
}
\caption{Ablation on memory consolidation on BrowseComp-Plus (LasJ).}
\label{tab:ablation_conso}
\end{table}

Without consolidation, Memory Shards grow monotonically as each newly produced gist is appended verbatim to the existing content. Over the course of a long-horizon task, this causes the gist loaded back into the Planner's context to become increasingly verbose and redundant. Two consequences follow. First, Planner PFLOPs rise notably (606 $\to$ 768, $+$26.5\%), since the Planner must attend over a longer and noisier memory context at each reasoning step---a direct manifestation of the quadratic attention cost that consolidation is designed to suppress. Second, the accumulated noise degrades reasoning quality: the Planner receives overlapping or conflicting statements across appended gists, making it harder to identify the most task-relevant signals, which leads to a 6.6 percentage point drop in LasJ (0.786 $\to$ 0.720).

These results highlight the dual role of consolidation in ActiveMem. Beyond simply reducing memory footprint, consolidation actively distills accumulated evidence into a compact, non-redundant form, ensuring that the Planner always operates over a clean and bounded context regardless of how many documents have been processed. This property is precisely what allows ActiveMem to maintain both high accuracy and low Planner cost as task length increases.

\subsubsection{Memory Hit Rate as a Search Saturation Indicator}
\label{appendix:cache_hit}

\begin{table}[h]
\centering
\small
\setlength{\tabcolsep}{4pt}
\begin{tabular}{lccc}
\toprule
\textbf{Step Range} & \textbf{Easy} & \textbf{Medium} & \textbf{Hard} \\
\midrule
1--3   & 3.1 & 2.0  & 3.8  \\
4--6   & 5.4 & 4.8  & 6.3  \\
7--10  & 7.4 & 6.8  & 7.7  \\
11--15 & 7.5 & 7.8  & 8.7  \\
16--20 & 6.7 & 9.3  & 8.4  \\
21--30 & 5.0 & 10.2 & 9.4  \\
31--50 & --- & 11.2 & 10.7 \\
\bottomrule
\end{tabular}
\caption{Memory hit rate (\%) by step range and difficulty level.}
\label{tab:cache_hit}
\end{table}

The three difficulty levels exhibit qualitatively different memory hit trajectories, as shown in Table~\ref{tab:cache_hit}.
For easy cases, the hit rate peaks at steps 7--10 (${\approx}7.5\%$) and subsequently declines to 5--6\%.
Easy cases average only 8.8 steps, so cases that reach step 16+ are a self-selected subset that failed to find an answer early.
At this point, having exhausted the small cluster of documents relevant to a straightforward question, the agent is forced to issue increasingly peripheral queries; these queries return documents far from the previously visited region, producing more first-seen documents rather than memory hits, and the hit rate consequently falls.
For medium cases, the hit rate rises monotonically from 2.0\% to 11.2\% with no reversal, reflecting an agent that has sufficient steps to saturate its local retrieval space while repeatedly circling nearby documents.
Hard cases also rise monotonically but tell a more nuanced story.
\begin{figure*}[t]
  \centering
  \includegraphics[width=\textwidth]{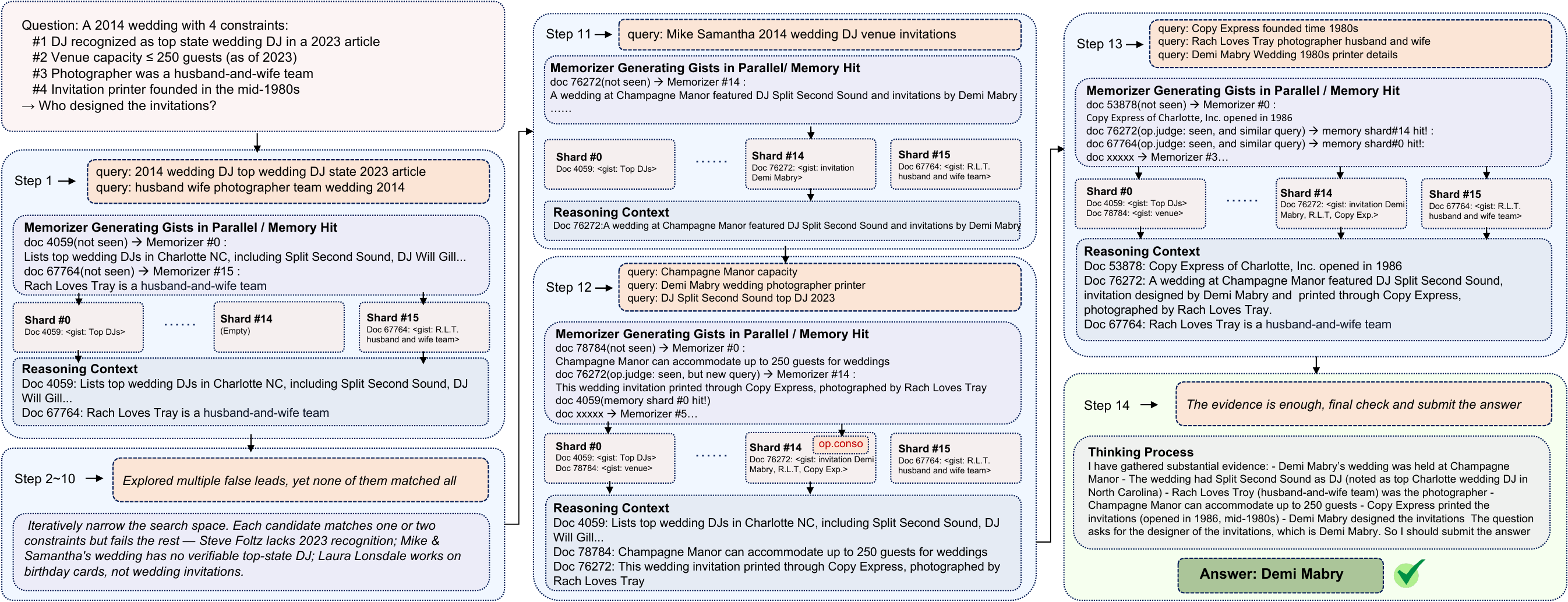}
  \vspace{6pt}
  \includegraphics[width=0.667\textwidth]{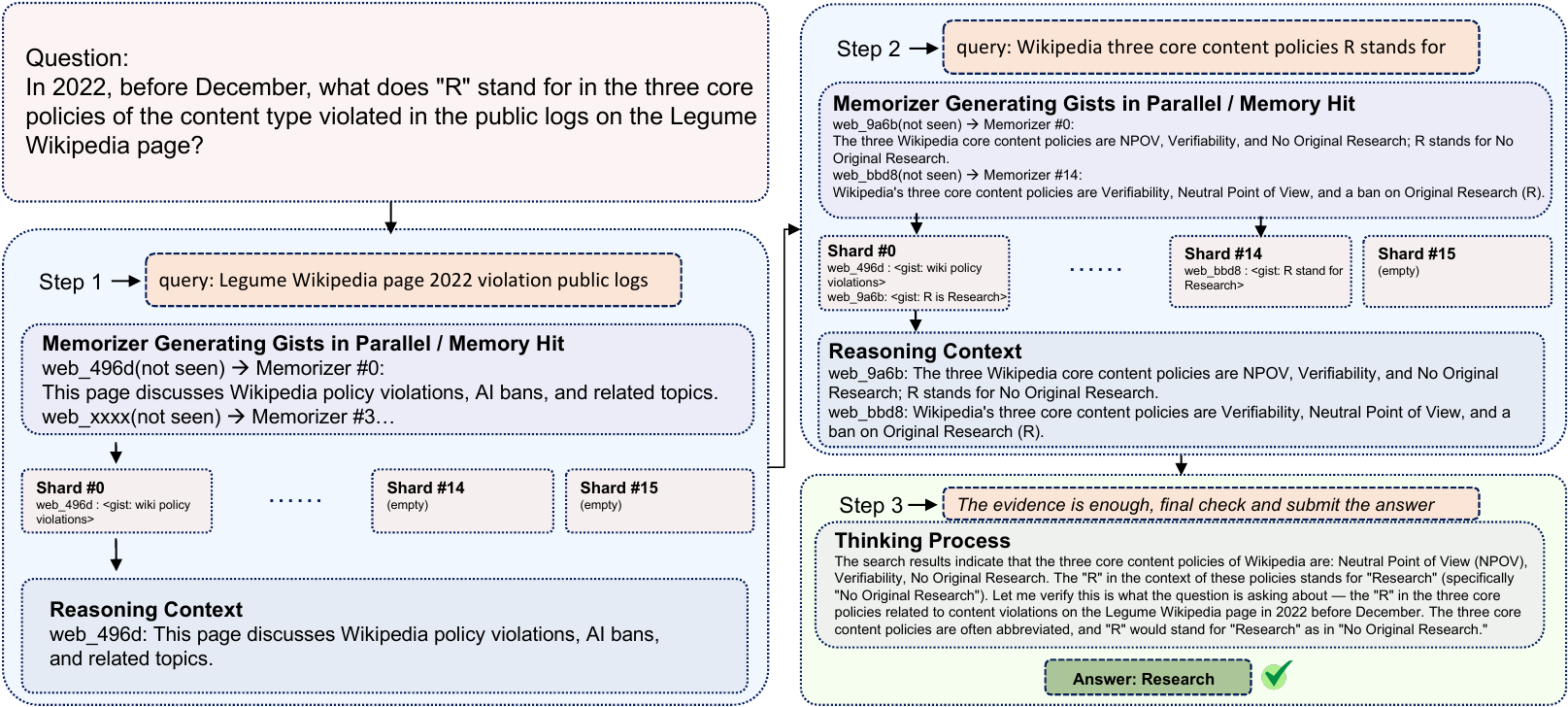}
  \caption{Case study traces for a BrowseComp-Plus instance \textbf{(top)} and a GAIA instance \textbf{(bottom)}. Each box shows the query issued, the Memorizer-produced gist or Memory Hit result, and the resulting Reasoning Context passed to the Planner. The BrowseComp-Plus question involves multiple interrelated constraints and requires a broader, longer search trajectory to satisfy all of them simultaneously. The GAIA question has fewer constraints and converges to the answer in far fewer steps.}
  \label{fig:case_studies}
\end{figure*}
The natural hypothesis is that harder questions cause the agent to fail to find key information, fall back on repeated searches, and thereby accumulate higher memory hits. While this mechanism is valid, Table~\ref{tab:cache_hit} reveals an important qualification: hard cases start slightly above medium at steps 1--10, yet fall below medium thereafter. This is because the inherent nature of hard questions prevents the agent from accumulating a dense cluster of retrieved documents around any single topic---the search space is too broad and the relevant signals too sparse for repeated queries to converge on the same small set of documents. As a result, hard-case agents operate within a wide but diffuse retrieval space, only occasionally re-encountering the same documents, which keeps the hit rate from rising as steeply as in medium cases where the agent circles more tightly around a narrower topic area.

These patterns together suggest that memory hit rate can serve as a practical search saturation indicator. A low hit rate in early steps reflects an effective exploration phase in which the agent is discovering new documents at each turn. A rising hit rate in later steps signals that the locally reachable document set is approaching exhaustion and the agent has begun cycling within already-visited content. When the per-step hit rate exceeds ${\approx}10$--$12\%$ for several consecutive steps, continued search is unlikely to yield further marginal benefit, and an early-stopping criterion could be triggered accordingly. We leave the formal design and evaluation of such a mechanism to future work.

\subsubsection{Case Study}
\label{appendix:case_study}

We present two representative cases from BrowseComp-Plus and GAIA to illustrate the working pipeline of ActiveMem. GAIA questions typically involve fewer and more directly verifiable constraints, allowing ActiveMem to converge to an answer within a small number of steps. BrowseComp-Plus questions are substantially harder: they involve multiple interrelated constraints that must be jointly satisfied, requiring longer reasoning trajectories and a larger number of retrieved documents before sufficient evidence accumulates.

\paragraph{BrowseComp-Plus.}
Figure~\ref{fig:case_studies} (top) traces the trajectory for a question asking for the designer of wedding invitations from a 2014 ceremony, subject to four constraints: the DJ was recognized as top state wedding DJ in a 2023 article, the venue accommodates up to 250 guests, the photographer was a husband-and-wife team, and the invitation printer was founded in the mid-1980s.

At Step~1, two parallel queries initialize two shards: doc4059 (Charlotte NC top DJ list including Split Second Sound) and doc67764 (a husband-and-wife photo team). Steps~2--10 explore multiple false leads---Steve Foltz lacks 2023 recognition; Mike \& Samantha's wedding has no verifiable top-state DJ; Laura Lonsdale works on birthday cards---and none satisfies all four constraints simultaneously. The pivot occurs at Step~11, when doc76272 is retrieved and distilled into: \textit{``A wedding at Champagne Manor featured DJ Split Second Sound and invitations by Demi Mabry,''} linking the DJ, venue, and answer candidate in one gist. At Step~12, the Similarity Judge returns \textit{SIMILAR} for doc4059, triggering a Memory Hit that confirms Split Second Sound's top-state recognition without re-invoking the Memorizer; for doc76272 the Judge returns \textit{NEW}, so the Memorizer re-distills and the Operator consolidates the result into Shard~\#14. At Step~13, a fresh retrieval confirms Copy Express opened in 1986, and a Memory Hit on doc67764 confirms Rach Loves Troy as the husband-and-wife photographer. All four constraints are satisfied, and the Planner submits \textit{Demi Mabry} at Step~14.

\paragraph{GAIA.}
Figure~\ref{fig:case_studies} (bottom) traces a two-hop question: what does ``R'' stand for in the three core policies of the content type violated in the Legume Wikipedia page logs before December 2022?

At Step~1, the query \textit{``Legume Wikipedia page 2022 violation public logs''} retrieves a page indicating that the violation concerned Wikipedia's core content policies, identifying the content type needed for the second hop. At Step~2, the query \textit{``Wikipedia three core content policies R stands for''} retrieves two independent documents that both identify ``R'' as ``Research'' (No Original Research). The Planner submits \textit{Research} after a brief verification pass, completing the trajectory in three steps with two fresh retrievals and no Memory Hits.

% ─────────────────────────────────────────────
\subsection{The Use of AI Assistants}
\label{appendix:ai_assistants}
% ─────────────────────────────────────────────

In this paper, AI assistants (ChatGPT and Claude) were used exclusively for language polishing, including grammar correction, phrasing, and stylistic refinement. They were not used to generate scientific content such as research ideas, methods, experiments, or related work. No confidential, personal, or proprietary information was shared with the models. The authors take full responsibility for the scientific content, which was entirely authored and verified by the authors.

\end{document}